\documentclass[sigconf]{acmart}

\AtBeginDocument{%
  }

\setcopyright{acmlicensed}
\copyrightyear{2018}
\acmYear{2018}
\acmDOI{XXXXXXX.XXXXXXX}

\acmConference[Conference acronym 'XX]{Make sure to enter the correct
  conference title from your rights confirmation emai}{June 03--05,
  2018}{Woodstock, NY}
\acmISBN{978-1-4503-XXXX-X/18/06}

\usepackage{lipsum}
\usepackage{xspace}
\usepackage{multicol}
\usepackage{multirow}
\usepackage{algorithm}
\usepackage{algpseudocode}
\usepackage{amsmath}
\usepackage{tcolorbox}
\usepackage{makecell}
\begin{document}

\title{Awaking the Slides: A Tuning-free and Knowledge-regulated AI Tutoring System via Language Model Coordination}

\author{Daniel Zhang-Li}
\authornote{Both authors contributed equally to this research.}
\email{zlnn23@mails.tsinghua.edu.cn}
\author{Zheyuan Zhang}
\authornotemark[1]
\email{zheyuan-22@mails.tsinghua.edu.cn}
\author{Jifan Yu}
\email{yujifan@tsinghua.edu.cn}
\affiliation{%
  \institution{Tsinghua University}
  \city{Beijing}
  \country{China}
}

\author{Joy Lim Jia Yin}
\email{lin-jy23@mails.tsinghua.edu.cn}
\author{Shangqing Tu}
\email{tsq22@mails.tsinghua.edu.cn}
\author{Linlu Gong}
\email{gll21@mails.tsinghua.edu.cn}
\author{Haohua Wang}
\email{wanghh24@mails.tsinghua.edu.cn}
\affiliation{%
  \institution{Tsinghua University}
  \city{Beijing}
  \country{China}
}

\author{Zhiyuan Liu}
\email{liuzy@tsinghua.edu.cn}
\author{Huiqin Liu}
\email{liuhq@tsinghua.edu.cn}
\author{Lei Hou}
\email{houlei@tsinghua.edu.cn}
\author{Juanzi Li}
\email{lijuanzi@tsinghua.edu.cn}
\affiliation{%
  \institution{Tsinghua University}
  \city{Beijing}
  \country{China}
}

\renewcommand{\shortauthors}{Trovato et al.}

\begin{abstract}
The vast pre-existing slides serve as rich and important materials to carry lecture knowledge.
However, effectively leveraging lecture slides to serve students is difficult due to the multi-modal nature of slide content and the heterogeneous teaching actions.
We study the problem of discovering effective designs that convert a slide into an interactive lecture.
We develop \model, a tuning-free and knowledge-regulated intelligent tutoring system that can
(1) effectively convert an input lecture slide into a structured teaching agenda consisting of a set of heterogeneous teaching actions;
(2) create and manage an interactive lecture that generates responsive interactions catering to student learning demands while regulating the interactions to follow teaching actions.
\model contains a complete pipeline for learners to obtain an interactive classroom experience to learn the slide.
For teachers and developers, \model enables customization to cater to personalized demands.
The evaluation rated by annotators and students shows that \model is effective in outperforming the remaining implementation.
\model's online deployment has made more than $200K$ interaction with students in the $3K$ lecture sessions.
We open source \model's implementation in \url{https://anonymous.4open.science/r/slide2lecture-4210/}.

\end{abstract}



\received{20 February 2007}
\received[revised]{12 March 2009}
\received[accepted]{5 June 2009}

\newcommand{\modelname}{Slide2Lecture}
\newcommand{\model}{\textit{\modelname}\xspace}
\newcommand{\EduSequence}{\mathcal{L}}
\newcommand{\edusequence}{\mathscr{l}}
\newcommand{\teachintention}{\mathscr{t}}

\maketitle
\section{Introduction}


The vast amount of learning materials available on the Internet, thanks to the sharing and dissemination of knowledge by educators, has enabled countless individual learners to achieve personal development through self-study~\cite{apperson2006impact}.
However, despite the availability of numerous course slides and textbooks, students often experience confusion during self-learning, which requires additional clarification to address these diverse learning needs caused by different knowledge backgrounds.
Without personalized guidance and explanations from teachers and assistants in traditional classrooms~\cite{bloom19842}, individuals cannot interact directly with educational materials, making self-study a difficult and challenging endeavor~\cite{hone2016exploring}.


\textbf{Existing Studies.} Despite a few previous efforts to build Intelligent Tutoring Systems (ITS) that provide self-learners with adaptive learning experiences~\cite{dan_educhat_2023,sonkar_class_2023,deng_towards_2023, tu2023littlemu}, these approaches often require significant investment and extensive model training.
With the rapid development of Large Language Models (LLMs), researchers have attempted to create ITS by abstracting complex teaching behaviors with LLM workflows~\cite{chen_empowering_2023, liu_scaffolding_2024}, which remains largely theoretical and challenging to implement without actual course materials. Nevertheless, constructing personalized ITS from specific teaching slides involves numerous technical challenges:


$\bullet$ \textbf{Multi-Modal and Structured Teaching Material Understanding.}
Online massive teaching materials encompass multi-modal~\cite{brock2011empowering} and hierarchically structured knowledge~\cite{atapattu2017comprehensive}, such as relationships between pages and sections.
To ensure that the explanations of the lecture slides are accurate, informative, and seamless, understanding the slides must cover multiple modalities\footnote{Textual and visual information.} and different levels of knowledge content\footnote{Page level and section level.}.
Therefore, designing systems that can fully leverage the abundance of knowledge for tutoring is a significant challenge worth exploring.

$\bullet$ \textbf{Heterogeneous Teaching Action Generation.} 
Previous researchers have focused on specific teaching actions such as explanations~\cite{sonkar_class_2023}, Q\&A~\cite{tu2023littlemu}, and exercises~\cite{cui2023adaptive}. However, generating and arranging appropriate heterogeneous teaching actions grounding on the slides, similar to experienced educators, continues to be a significant hurdle in a comprehensive tutoring system to assist learners effectively.
For example, determining when to test learners with what exercises poses strict requirements on the planning and generation capabilities of the tutoring system.

$\bullet$ \textbf{Interactive Tutoring Management.}
Because of the diverse knowledge backgrounds and study habits of different learners~\cite{karuovic2021students}, following the personalized learning pace while keeping learners engaged in tutoring environments is especially important~\cite{pursel2016understanding}.
Therefore, how to dynamically understand the learners' intentions and learning states, and effectively use teaching actions combined with real-time Q\&A and discussions, present an intricate challenge for interactive tutoring managements in the system.

\begin{figure*}[t]
    \centering
    \includegraphics[width=1\linewidth]{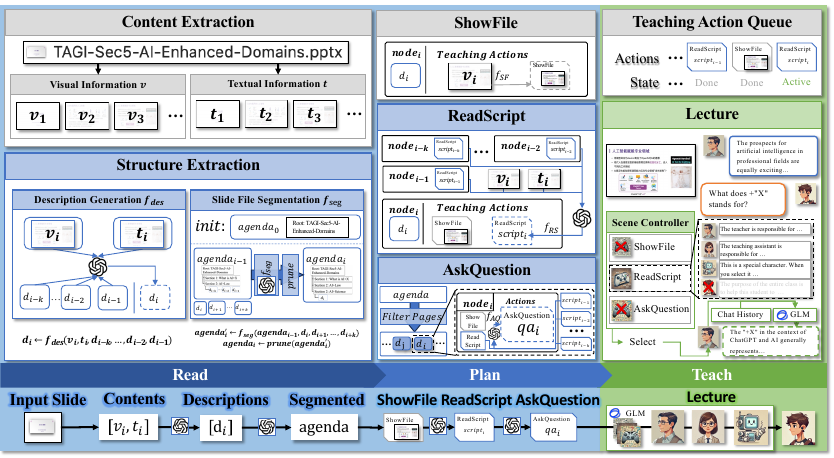}
    \caption{Framework overview for \model. A slide file is processed to extract its contents and structure. The \textit{Plan} subsystem then takes the extracted materials to generate a set of teaching actions. Finally, the \textit{Teach} module calls the matching scene controller to control the agents, creating an interactive lecturing to serve user learning while aligning with slide.}
    \label{fig:framework}
    \Description[]{}
\end{figure*}


\begin{algorithm}[t]
\caption{Pre-Class Subsystems: \textit{Read} and \textit{Plan}.}
\begin{algorithmic}[1]
\small
\Require $\mathcal{F}, f_{des}, f_{seg}, prune, f_{SF}, f_{RS}, f_{AQ}$
\State $Content \gets []$, $agenda_0 \gets \mathcal{F}.title$
\For{$i, p \in enumerate(\mathcal{F})$}
    \State $v_i \gets p.visual$, $t_i \gets p.textual$ \Comment{Content Extraction}
    \State $Content.append(\{v_i,t_i\})$
    \State $d_i \gets f_{des}(v_i,t_i,d_{i-k},\dots,d_{i-2},d_{i-1})$ \Comment{Description Generation}
\EndFor
\For{$i=1,2,\ldots,len(Content)$} \Comment{Slide File Segmentation}
    \State $agenda_i \leftarrow prune(f_{seg}(agenda_{i-1},d_i,d_{i+1},\dots,d_{i+k}))$
\EndFor
\For{$node \in  agenda.get\_leaf\_nodes()$}
    \State $p_i \gets node.get\_corresponding\_page()$ \Comment{$ShowFile$ Generation}
    \State $(ShowFile,p_i.id)\gets f_{SF}(p_i)$
    \State $node.teaching\_actions.append(ShowFile,p_i.id)$
    \State $v_i, t_i \gets node.get\_content()$ \Comment{$ReadScript$ Generation}
    \State $(ReadScript, script_i)\gets f_{RS}(v_i,t_i,script_{i-k},\ldots,script_{i-1})$
    \State $node.teaching\_actions.append(ReadScript, script_i)$
\EndFor
\For{$section \in agenda.sections$}
    \If{$len(section.flatten()) >= k$}
        \State $node\gets section.last\_node$ \Comment{$AskQuestion$ Generation}
        \State $i \gets node.page\_index()$
        \State $(AskQuestion, qa_i) \leftarrow f_{AQ}(script_i,\dots,script_{i-k}])$
        \State $node.teaching\_actions.append(AskQuestion, qa_i)$
    \EndIf
\EndFor
\end{algorithmic}
\label{algo:pre-class}
\end{algorithm}
\textbf{Present Work.} To address these crucial and complex requirements, we introduce \model, an innovative framework for a knowledge-regulated intelligent tutoring system designed and evaluated jointly by a group of engineers, computer researchers, and education experts.
In particular, inspired by the pattern of human tutors, we formulate the tutoring process into three stages, \textbf{\textit{Read}-\textit{Plan}-\textit{Teach}}, where subsystems are implemented for each stage to tackle the challenges accordingly.
(1) The \textit{Read} subsystem processes the comprehensive input file into a unified representation, including knowledge of both content (visual and textual) and structure.
(2) The extracted slide is then handed over to the \textit{Plan} subsystem, where the extracted knowledge is formalized into various types of teaching actions that can be generated.
(3) In contrast to the \textit{Read} and \textit{Plan} subsystems which together serve for pre-class generation, the \textit{Teach} subsystem takes the list of teaching actions as inputs and manages the lecture for interactive tutoring regulated by the knowledge within the current active teaching action.

\textbf{Evaluation.} We test the performance of \model via experimental evaluation and online deployment.
Results from the experiments demonstrated the effectiveness of both lecture generation to match the slide and the delivery of lectures for knowledge learning.
Feedback from online deployments, in advance, illustrated the benefits of using such systems in future education, where students favored \model's free and easy learning experience.

\textbf{Impact and Beneficial Groups.} As a system with online deployment used by $556$ students for two experimental courses, \model aims to serve the extensive ITS community, potentially benefiting various stakeholders:
(1) \textit{Learners.} We provide an interactive classroom experience, allowing quick generation and participation in lectures based on provided slides.
(2) \textit{Teachers.} The system features user-friendly visual interfaces, where teachers can not only upload slides to serve lectures, but also practice lecturing from the perspective of the student.
(3) \textit{Computer Scientists and Developers.} Core functionalities are offered in modular implementations with thorough documentation, facilitating detailed review and modification with minimal effort.
(4) \textit{Academics for Education.} The system records interactions effectively, supporting analysis and observations.







\section{Preliminaries}
In this section, we first introduce the background techniques for building the proposed system and then formalize the task.
\subsection{Background Techniques}
The construction of an interactive and knowledge-regulated intelligent tutoring system (ITS) involves joint collaboration of intelligent tutoring techniques and powerful large language models.

\paragraph{\textbf{Intelligent Tutoring System}}
The investigation into the architecture of intelligent tutoring systems commenced during the initial era of computer-assisted instruction (CAI)~\cite{carbonell1970ai,alkhatlan_intelligent_2019, castro2021intelligent, tu2023littlemu}. The advent of contemporary machine learning methodologies has catalyzed recent endeavors to develop systems that are more interactive, personalized, emotionally supportive, and tailored to individual learning styles~\cite{mallik_proactive_2023, zhang2024simulating}. The introduction of ChatGPT~\cite{ouyang2022training} has had a profound impact on the educational domain~\cite{alqahtani2023emergent}. In particular, some researchers have discovered that, with adequate training, language models are capable of effectively addressing student inquiries~\cite{baladon2023retuyt, dan_educhat_2023}. Furthermore, integrating language models with graph-based techniques has demonstrated substantial promise in mapping learning trajectories for specific concepts~\cite{deng_towards_2023}.

\paragraph{\textbf{Language Model Coordination}}
Large Language Models (LLMs) demonstrate advanced functionalities that have profoundly impacted various domains~\cite{bubeck2023sparks}, particularly in increasing interactivity within intelligent tutoring systems. Owing to ongoing discoveries in multi-agent and tool-utilization studies~\cite{Park2023GenerativeAgents, qian2023communicative, schick2024toolformer}, novel planning and interaction techniques for ITS have been devised. A recent investigation~\cite{chen_empowering_2023} has also revealed the potential of LLMs to autonomously design and deliver lectures by integrating diverse functionalities into multiple tools. Subsequently, MWPTutor~\cite{chowdhury_autotutor_2024} examines the coordination with LLMs to facilitate the instruction of students in specific mathematical word problems. These insights have propelled our research into the development of an LLM-based ITS, poised to establish a foundational framework for lecture-level and knowledge-regulated intelligent tutoring systems, encouraging the role of LLMs in future educational paradigms.

\subsection{Problem Formulation}

\paragraph{Definition 1.} \textbf{Lecture Slide}, denoted as $\mathcal{F}$, is a file rich in diverse form of knowledge ($\teachintention_{1\sim n}$) that could be used for teaching.

\paragraph{Definition 2.} \textbf{Intelligent Tutoring System} is a system that takes the given interaction history, $\mathcal{H}^t$, to generate a set of interactions, $S_t$, in response to user query, $U_t$, where $t$ is the count of user queries.

\paragraph{Problem 1} \textbf{Slide Knowledge-regulated Intelligent Tutoring}
\textit{For a given lecture slide $\mathcal{F}$, the task is to create a set of interactions $S_t$ in response to the last user query $U_t$, while ensuring that the interaction history $\mathcal{H}^t$ is within the scope of knowledge of slide file $\teachintention_{1\sim n}$}.




\section{\modelname}

In this section, we first overview the design principle and the workflow of \model, and then elaborate on its implementation.










\subsection{Framework Overview}
\label{sec:method-framework}
Figure~\ref{fig:framework} shows the overview of \model, which consists of three subsystems, \textit{Read}, \textit{Plan}, and \textit{Teach}.
The \textit{Read} and \textit{Plan} jointly serve during pre-class (Algorithm~\ref{algo:pre-class}) and the \textit{Teach} subsystem is responsible for providing real-time in-class interactions.
The detailed workflow of these three subsystems is illustrated as follows:

(1) \textit{Read: Multi-Modal Knowledge Extraction.} 
This subsystem is designed to extract rich and diverse knowledge within the slide.
We aim to retain the hierarchical structure of the course slides. Leveraging the advanced capability to understand multi-modal inputs of modern LLMs~\cite{yang2023dawn, team2023gemini}, this subsystem not only extracts the visual and textual contents of the slide but also consists of a pipeline to rebuild a \textbf{tree-formed agenda} to extract the structure knowledge.
These outputs serve as uniform input knowledge to augment and regulate the generation of later subsystems.
In addition, this design also potentially enables developers to transform the implementation to input materials other than lecture slides.

(2) \textit{Plan: Heterogeneous Teaching Action Generation.}
This subsystem is implemented to formalize the extracted knowledge into a set of heterogeneous \textbf{teaching actions}, such as \textit{ShowFile}, \textit{ReadScript}, and \textit{AskQuestions}, which will be practiced during the lecture.
The \textit{Plan} subsystem includes a modular implementation to handle each type of teaching action and will output a serialized action queue, which can be further used in the subsequent teaching process. 
The generated teaching action queue allows for a high degree of flexibility for adjustments. By formalizing the teaching actions before class, human teachers are allowed to revise the teaching actions.

(3) \textit{Teach: Knowledge-regulated Interactive Tutoring.} 
This subsystem is responsible for creating an interactive lecture for the student while ensuring that the interactions are regulated by long-formed teaching actions.
This requirement is ensured by the collaborative work of \textbf{scene controllers} and \textbf{interacting agents}.
Different scene controllers are created to optimize and implement diverse process controls for each teaching action by selecting and controlling agents. Each agent is assigned to handle a unique form of interaction (e.g., the teaching assistant is responsible for answering questions and maintaining discipline). The responsibilities of each agent are provided to the scene controller in the form of role descriptions. Therefore,
removing a form of interaction can be done simply by removing the role description for the agent, preventing the need to revise the complex agent implementation.

With the modularized design of the teaching activities during both \textit{Plan} and \textit{Teach} stages, scholars and practitioners can easily add, customize, and refine the implementation of individual teaching activities to align with specific pedagogical requirements.

In the remainder of this section, we explain the technical details of \model through an implementation used for two courses taken by $556$ students. We then conduct an independent set of comprehensive analyses in the next section for evaluation.




\subsection{Read: Multi-Modal Knowledge Extraction}
\label{sec:method-pre-class-read}

The function of multi-modal knowledge extraction is to prepare the multi-modal long-formed slides for generating teaching actions.
\model extracts both content and structure knowledge from the given slide file to simplify and unify the input.

\subsubsection{\textbf{Content Knowledge}}
To extract information from the given slide, we focus primarily on extracting two types of content: textual ($t$) and visual ($v$).
For the textual content, we use $python-pptx$ to extract all the texts within each slide page.
For visual content, we convert each slide page into an image using $libreoffice$ and $pymupdf$.
The content knowledge for each slide is organized into textual and visual information for further processing.


\subsubsection{\textbf{Structure Knowledge}} In addition to the content knowledge, we further organized the slides into a tree-formed structure. 
The structured knowledge is extracted in an iterative manner, dividing the input slide into sections based on the content knowledge of each page, where each section contains multiple pages ($\{p_i, p_{i+1}, \dots, p_j\}$).
This extraction process aims to segment the slide into a set of sequences, where each sequence is continuous and represents a section.
The extract structure knowledge offers two main advantages. First, it helps preserve the section and ordering of the slides, ensuring that knowledge points are positioned within the overall lecture, and maintaining the hierarchy of knowledge, such as prerequisite relationships. Second, the tree-formed structure benefits the process of teaching action generation. For example, questions and exercises can be inserted at the end of each section to facilitate student learning. In \model, the structure knowledge of the slide is extracted using two components: Description Generation and Slide File Segmentation.

\textbf{Description Generation ($f_{des}$).} To abstract the content and purpose of each page based on its content knowledge, while also reducing the content length to aid the subsequent tree-formed structure generation process, we implemented a tool that generates the \textbf{description} for each slide page as a shorter replacement for the original content.
In detail, this module takes the content of the page, $\{v_i,t_i\}=p_i$, and the descriptions of the previous pages, $\{d_{i-k},\dots,d_{i-2},d_{i-1}\}$, to iteratively generate the description.

$$
d_i \leftarrow f_{des}(v_i,t_i,d_{i-k},\dots,d_{i-2},d_{i-1}) 
$$

\begin{figure*}[t]
    \tiny
    \centering
    \includegraphics[width=0.33\linewidth]{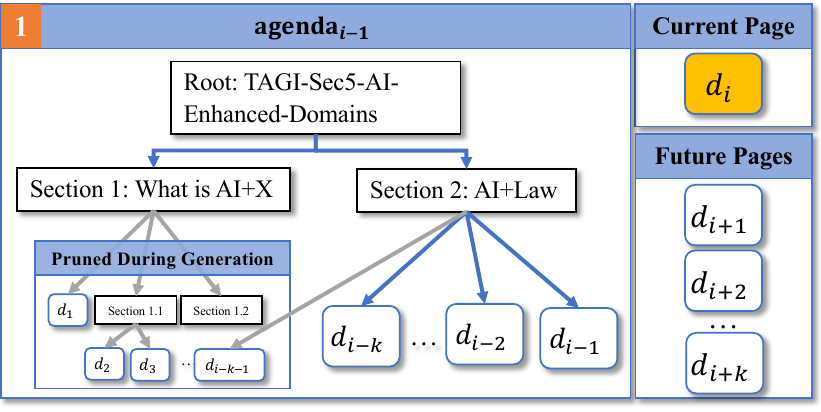}
    \includegraphics[width=0.33\linewidth]{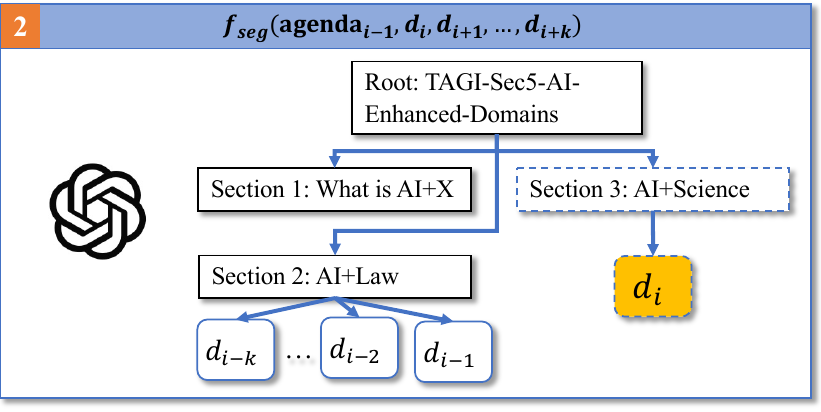}
    \includegraphics[width=0.33\linewidth]{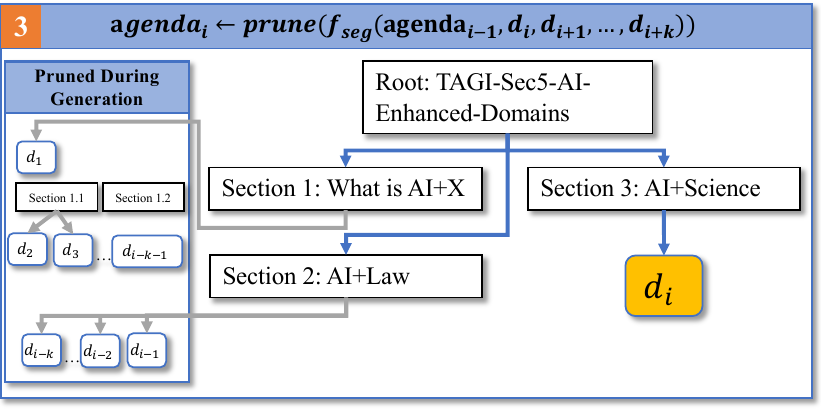}
    \caption{Example of slide file segmentation ($f_{seg}$ and $prune$).
    In each iteration, the LLM is asked to insert a page into the existing agenda. We then prune the nodes that are not direct siblings of the ancestors to limit the context length.
    }
    \Description[]{}
    \label{fig:structure}
\end{figure*}

\textbf{Slide File Segmentation ($f_{seg}$).} This step is specially implemented to segment the long list of input pages within the slide by structuring these slide pages into a tree-form.
Since the focus for each page could range from the general topic of the entire course and the introduction to a specific concept, directly developing a method to produce a segmentation in the form of sequences is a nontrivial task.
Modern LLMs have the potential to perform comprehensive text understanding and knowledge-intensive tasks~\cite{yang2024harnessing}. However, they are limited by context length, constraining such powerful models from gaining sight of the entire file's structure. 

As shown in Figure~\ref{fig:structure}, instead of directly outputting the segmented sequences, we ask the LLM to output in the form of a tree, named $agenda$. In this structure, each page is initialized as a leaf node on the organized agenda, with the remaining nodes representing section names. To include contextual knowledge that is excluded from description generation for upcoming pages, during each step, $f_{seg}$ takes the description generated for the current page, $d_i$, the revised agenda from the previous step, $agenda_{i-1}$, and several future pages, $d_{i+1},d_{i+2},\dots,d_{i+k}$, to gain context for outputting the revised $agenda_i$.
Specifically, the output agenda $agenda_i$ is a revision of $agenda_{i-1}$, with the page $p_i$ represented as $d_i$ being a new leaf node and optional new section names as the nearest ancestry nodes before finding an existing ancestry. We initialize $agenda_0$ with the title of the slides configured by the teacher.

The resulting tree has a space complexity of at least $O(len(\mathcal{F}))$, which may exceed the LLM's context limit. To reduce complexity, after updating the generated content into the tree stored in memory, we prune the tree by removing unnecessary nodes that are not (1) ancestors of the newly appended leaf node or (2) directory siblings of these ancestors. The pruning helps to reduce context length to store previous sections that are folded in future generations.
$$
agenda_i \leftarrow prune(f_{seg}(agenda_{i-1},d_i,d_{i+1},\dots,d_{i+k}))
$$


\subsection{Plan: Heterogeneous Teaching Action Generation.}

The objective of teaching action generation is to take the extracted knowledge of the slide pages and generate a set of teaching actions regulated by this knowledge that can be further used in the interactive teaching part.
Our design focuses on two major concerns: (1) \textit{How should teaching actions be represented to capture the diverse teaching activities?} (2) \textit{How should heterogeneous teaching actions be generated within the model context?}

\subsubsection{\textbf{Teaching Action Representation}} 
The teaching activities (e.g. lecturing, giving quizzes) are abstracted into \textit{teaching actions} in \model. Specifically, teaching actions ($\teachintention$) are represented as $\teachintention = (type, value)$, where $type$ denotes the category of teaching action (e.g. $ShowFile$, $ReadScript$, and $AskQuestion$), and $value$ specifies the content of the action (e.g. the script to be read). This design allows classroom actions to be configurable, enabling developers and educators to add custom teaching actions as needed and seamlessly integrate them into the teaching process.


\subsubsection{\textbf{Teaching Action Generation}} As the teaching actions are related to specific slide pages, we generate and store teaching actions as lists within the agenda nodes of the slide pages, represented as $node.actions = [\teachintention_1, \teachintention_2, \dots]$. In \model, we provide the implementation of the following three key teaching actions:


(1) \textbf{ShowFile ($f_{SF}$)} is used to show the pages of the slides to the students in order. In \model, the function $ShowFile$ is generated first. For simplicity, we loop over the page directly in each sequence and use the page id as the value for the action, denoted as $(ShowFile,p_i.id)$. Therefore, when \model performs $ShowFile$, the next page of the slides will be displayed.

(2) \textbf{ReadScript ($f_{RS}$)} is involved in introducing the content within each slide page.
For each slide page, we generate a corresponding teaching script. We aim to ensure that the scripts cover the details of the respective slide page and maintain a consistent tone across different pages. Therefore, we (1) recall the original visual ($v_i$) and textual ($t_i$) information extracted from the source slide page ($p_i$), (2) supplement it with the scripts generated for the previous k pages, and (3) concatenate this information and prompt the LLM to generate a script in the teacher's tone. The generated $ReadScript$ action is denoted as $(ReadScript, script_i)$.
The script will be used in the future by the agent teacher to introduce the knowledge described in the slide page.

(3) \textbf{AskQuestion ($f_{AQ}$)} is used to instruct the teacher agent to give quizzes to the student and provide explanations in response to the student's solution.
We leverage the structure of the agenda and generate questions at the end of the sections with at least $k$ slide pages to ensure that questions are generated only for sections with sufficient content.
LLMs are provided with the current script ($script_i$) as the direct source to generate questions and corresponding answers. The recently generated scripts from previous slide pages ($[script_{i-k},\dots, script_{i-1}]$) are also included as additional support to improve the quality of questions. The $AskQuestion$ action is denoted as $(AskQuestion, qa_i)$.
We consider both $single-choice$ and $multiple-choice$ questions in the current system, which leads to the value of the $AskQuestion$ action to contain four variables, $qa_i=(question, question\_type, options, answer)$.

The generated $qa_i$ will be appended to the action list, indicating that at the end of the learning process of each section, the teacher will ask questions and provide explanations after the student made a prediction to strengthen understanding.


\subsection{Teach: Knowledge-regulated Interactive Tutoring}
\label{sec:method-inclass}

The \textit{Teach} subsystem is responsible for providing interactive tutoring. It generates responses and manages an interactive lecture for the student user's queries ($U$) while ensuring the flow of the lecture is aligned with each teaching action ($\teachintention$). This subsystem is composed of two modules: Interacting Agents and Scene Controller.

\subsubsection{\textbf{Interacting Agents}} 
\model considers three types of interactions in the classroom: (1) giving lectures and providing explanations to the student user; (2) ensuring that the lecture follows the safety guidelines and handles unsafe inputs;
(3) providing other interactions that are important for the lecture process. Three LLM-based agents are implemented for each interaction:
For the first type of interaction, we iconize a \textbf{teacher agent} to deliver tutoring and answer user's questions.
For the second type of interaction, the \textbf{teaching assistant agent} is crafted and optimized to manage the discipline of the classroom.
To determine the termination of the teaching action and iconize other forms of interactions, \model also includes the \textbf{system agent} to control the system and determine the termination of the teaching actions.
The system agent is unaware to users, but serves crucial interactions, such as displaying a slide page and terminating a teaching action.

\subsubsection{\textbf{Scene Controller}}
Following previous works that utilize a managing agent to control the multi-agent scenarios~\cite{wu2023autogen, yue2024mathvc, zhang2024simulating}, we introduce the Scene Controller to select the appropriate agent to handle the lecture.
To create diverse interaction experiences for different actions, the behavior of the Scene Controller will vary from different teaching actions, 
which is described below:

(1) \textbf{ShowFile Controller} controls the system agent to change the displayed slide page followed by the termination of the action.

(2) \textbf{ReadScript Controller} would first control the teacher agent to provide the script to the user.
Then, if the user chooses to interact with the system, the controller would iteratively select the appropriate agent to generate a response, until it selects the user or the system agent to terminate an action.

(3) \textbf{AskQuestion Controller} would first call the teacher agent to post the question. Then, it would wait for the user's response. Once the user submits a solution, the controller will provide the teacher agent with the correct answer and ask it to generate a response to provide an explanation to the student.

\begin{table}[th]
    \centering
    \small
    \caption{Evaluation result of $ReadScript$ generation.}
    \begin{tabular}{l|cccc|c}
        \toprule
        \textbf{Setting} & \textbf{Tone} & \textbf{Clarity} & \textbf{Supportive} & \textbf{Matching} & \textbf{Overall}\\
        \midrule
        S2T~\cite{nguyen_automatic_nodate} & 3.88 & 3.93 & 3.23 & 3.63 & 3.67\\
        SCP~\cite{olney_automatic_2024} & \textbf{4.03} & \underline{4.24} & 3.38 & 3.93 & \underline{3.90}\\
        \midrule
        \model & 4.00 & \textbf{4.25} & \textbf{3.57} & \textbf{4.18} & \textbf{4.00}\\
        \ w/o visual & 3.78 & 3.73 & \underline{3.44} & 3.51 & 3.61\\
        \ w/o context & 3.97 & 4.00 & 3.38 & \underline{4.03} & 3.84\\
        \midrule
        Human & \underline{4.02} & 4.07 & 3.38 & 3.98 & 3.86\\
        \bottomrule
    \end{tabular}
    \label{tab:teacher-readscript}
\end{table}



\section{Experiments}
In this section, we design experiments to analyze the characteristics of \model.
To test our design, we first include offline evaluation and user studies to evaluate the overall performance of the pre-class subsystems (Section~\ref{sec:exp-pre-class}) and in-class subsystem (Section~\ref{sec:exp-inclass}).
We then conduct detailed observations of the components to study the performance of the system in detail (Section~\ref{sec:exp-component}).
Finally, we deployed \model as an online learning platform at a university, where $556$ student users participated. We demonstrate the statistical results of these users on the platform (Section~\ref{sec:exp-live}).

\paragraph{\textbf{Implementation}}
We set the value k as $3$ (number of previous pages in Section~\ref{sec:method-pre-class-read}) for all experiments.
For pre-class generation, we employ GPT-4V as the LLM for course content generation.
For real-time interactions during in-class sessions, in accordance with the ethical considerations of the partnering university and the existing collaboration agreement, we utilized GLM4~\cite{glm2024chatglm}.
We input a maximum of $12$ utterances of chat history during in-class generation.
Most hyperparameters are set to default, but max\_tokens was set to $4,096$ for GPT-4V (Appendix~\ref{sec:appendix-implementation-hyperparameter}).
In Appendix~\ref{sec:appendix-implementation-prompt}, we also provide the system prompts to reproduce our work.

\begin{figure}[]
    \centering
    \includegraphics[width=1\linewidth]{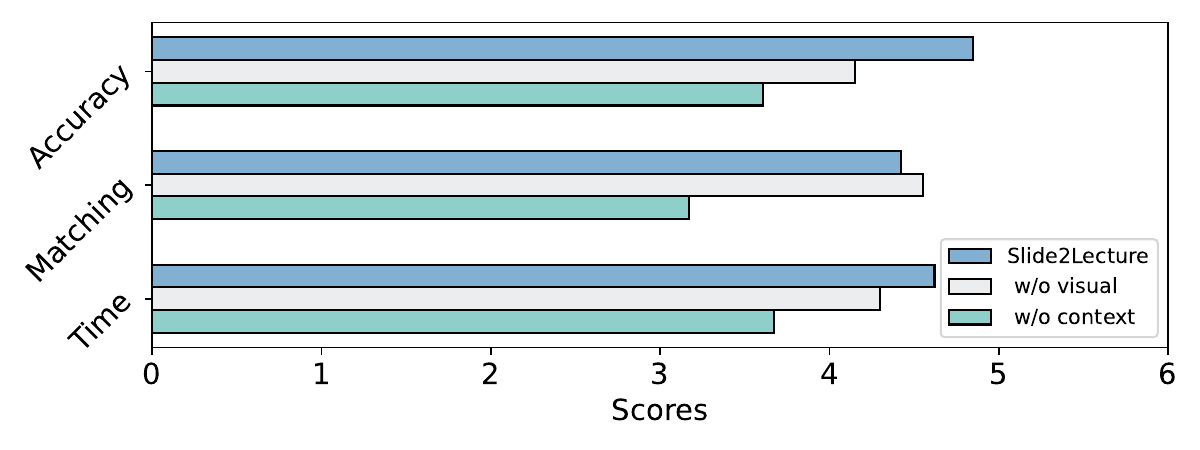}
    \caption{Evaluation result of $AskQuestion$ generation.}
    \label{fig:teacher-askquestion}
    \Description[]{}
\end{figure}
\begin{figure*}[thbp!]
    \centering
    \includegraphics[width=1\linewidth]{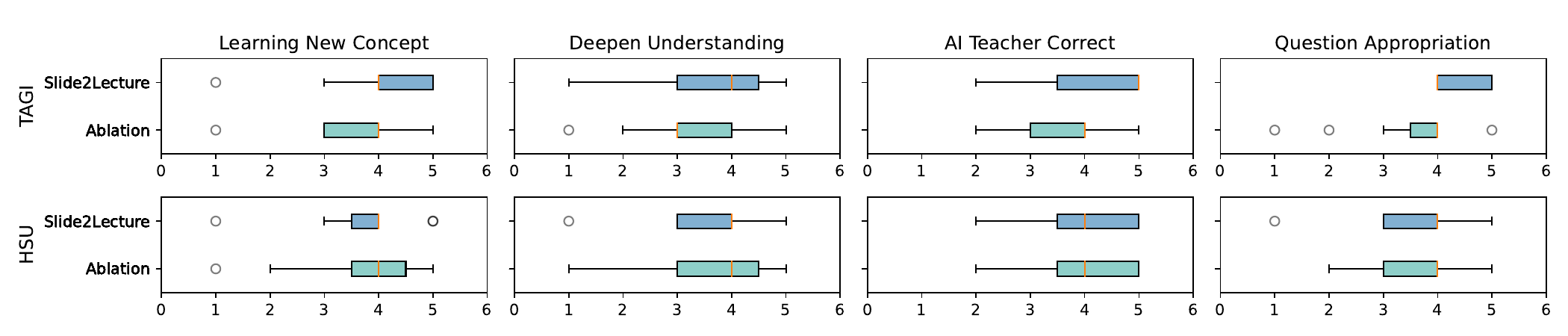}
    \caption{
        Results for in-class user study. The users are asked to rate the system after taking a lecture.
        We present the results in box-plot to display the median (red lines), 25\% to 75\% range ($IQR$ boxes), and outliers (dots) filtered with whiskers($1.5\times IQR$).
    }
    \Description[]{}
    \label{fig:user-study}
\end{figure*}
\subsection{Pre-Class Stages Evaluation}
\label{sec:exp-pre-class}
We perform offline evaluations to demonstrate the effectiveness of the method for the pre-class subsystems.
With the approval of the corresponding teachers, we include lecture slides drawn from the course \textit{Towards Artificial General Intelligence} (TAGI) and \textit{How to Study in University} (HSU) to conduct offline experiments.
The two slides contain $46$ and $61$ pages, respectively.
In the following paragraphs, we employ crowd-source annotators to assess the numerous \textit{ReadScript} actions (Section~\ref{sec:exp-offline-readscript}).
Subsequently, we engage experts to evaluate the less frequent \textit{AskQuestion} actions (Sec~\ref{sec:exp-offline-askquestion}).

\subsubsection{\textbf{ReadScript}}
\label{sec:exp-offline-readscript}
We first include two baseline setups to compare with \model implementation:
(1) We use the original instructions to reproduce \textit{\textbf{S}cript\textbf{2}\textbf{T}ranscript}~\cite{nguyen_automatic_nodate}, denoted as \textit{S2T}, which uses the titles to provide global information about the slide.
(2) We then reproduce \textit{\textbf{S}elf-\textbf{C}ritique \textbf{P}rompting}~\cite{olney_automatic_2024}, noted by \textit{SCP}, which includes a self-critique and refinement process to generate scripts.
Since \textit{SCP} uses the GPT-4~\cite{openai_gpt-4_2024} in its original literature, we include visual input to ensure that the performance is fully reproduced.
In contrast, we exclude visual inputs for \textit{S2T} as its original paper uses LLaMA~\cite{touvron_llama_2023}, which does not support out-of-box visual input.
To examine the influence of including visual and contextual (content in neighboring pages) information in the system, we also include two additional ablation setups, where the visual and contextual inputs are removed from \model's default implementation.
We use GPT-4V for all setups to ensure that the method performance is not affected by the capability of the backbone model.
We further include the scripts revised by the teachers and the TAs based on the slides as a human expert baseline.

\textbf{Metrics.}
We include $4$ different metrics during the evaluation of the generated scripts, each with a 5-point Likert scale, where $1$ stands for unacceptable and $5$ for optimal:
(1) \textit{Tone} stands for whether the script adopts the appropriate tone of a teacher delivering a lesson.
(2) \textit{Clarity} indicates whether the script is clear and easy to understand;
(3) \textit{Supportive} measures whether the script includes emotional support;
(4) \textit{Matching} grades how much the script matches the content of the slide page.
We compute the overall performance by averaging the scores for all metrics.

We run the pre-class pipeline for each baseline and ask the annotators to rate the scripts. To reduce bias between annotators, we ensure that each slide is labeled with $3$ annotators, where each of them is required to provide ratings for all settings of the same slide.

\textbf{Results.} As shown in Table~\ref{tab:teacher-readscript}, \model achieves the highest overall score of $4.00$, surpassing all baselines.
We observe that: (1) Visual input is important for script generation, as both \textit{S2T} and \model without visual inputs obtain poor matching scores;
(2) Contextual information is crucial for the quality of the scripts. Contents of previous pages not only improve the Clarity of the current page but also the Supportive and Matching, as they provide the model with coherent information.
(3) Surprisingly, our method slightly outperformed the human baseline across three dimensions. As LLMs have demonstrated a strong capability to follow instructions, they are inclined to strictly follow the slide contents with a more encouraging tone for the users. In contrast, human instructors tended to diverge more from the content, incorporating their own style and expanding on topics more freely.

\subsubsection{\textbf{AskQuestion}}
\label{sec:exp-offline-askquestion}
We tested the performance of \model in course question generation, by evaluating the quality and timing of the questions generated based on the lecture slides, considering that both of them are crucial in an open classroom setting.
Since preliminary studies have not focused on question-generation techniques with teaching slides and scripts, this evaluation focuses on observing performance change in ablation settings, where the visual and contextual inputs are removed respectively.
As the frequency of asking questions in class is influenced by individual teacher preferences, we omit human teacher evaluations of the questions.
We generate 3 questions per call to maintain an adequate sample size.

\textbf{Metric.}
Similar to the evaluation for \textit{ReadScript}, the human annotator is asked to rank each question from $1$ to $5$.
This ranking process involves three metrics:
(1) The \textit{Time} metric refers to whether it would be an appropriate time to ask a question;
(2) The \textit{Matching} metric is used to record whether the question matches pre-existing contexts;
(3) The \textit{Accuracy} metric is finally included to evaluate whether the question-answer pair is correct.

\textbf{Results.}
Figure~\ref{fig:teacher-askquestion} shows the result of the actions generated.
It is evident that (1) visual inputs contribute to better performance in producing the $AskQuestion$ actions, as the accuracy of the generated questions noticeably decreased when the LLM's access to visual input was restricted.
This decline is due to the fact that the slides provide supplementary knowledge that supports the generation process, which becomes limited and reliant only on the information from the page and its adjacent content. 
(2) We also observed larger drops when contextual inputs were removed.
This is likely because, in the slides used for the experiment, a significant amount of text content is effectively extracted, thus removing the contextual input leads to a greater loss of information.
This phenomenon caused by reduced input negatively impacted the no-context setting's performance in general, with significant declines evident across all three evaluation metrics.
(3) It's important to note that the absence of visual elements seemed to enhance the alignment of questions with the slide and scripts, which is due to the fact that this setting produced significantly fewer questions and was less effective at identifying locations for question generation.
We will discuss this observation further during the analysis of the segmentation component (Section~\ref{sec:exp-component-segmentation}). We demonstrate some question cases in the Appendix~\ref{sec:appendix-example-case}.

\subsection{In-Class Stage Evaluation}
\label{sec:exp-inclass}
We conduct a series of user studies to evaluate the design for the in-class \textit{Teach} subsystem. Specifically, our human evaluation team consists of $22$ students. In the experiment, all students are asked to use the automatically generated lecture plans from Section~\ref{sec:exp-pre-class} to assess the system.
Each student is required to interact freely with the system and go through at least $40$ pages of each lecture.

		

We also conduct a parallel experiment as the ablation study during this evaluation. To replicate the traditional meta-agent strategy, we eliminate the role descriptions of the system agent and omit the prompt injection for the teacher while generating responses. This setting can help us to observe whether the design of the \textit{Teach} module is effective from the user's perspective.

\textbf{Metric.}
Students are required to evaluate the system by assigning several $1$ to $5$ scores for each lecture, and the metrics include:
(1) \textit{Learning New Concepts}: whether the system can effectively introduce and teach new concepts to students; 
(2) \textit{Deepen Understanding}: how in-depths can the system reach when teaching concepts; (3) \textit{AI Teacher Correct}: the ability of the teacher agent to provide correct explanation after the student has provided a solution; (4) \textit{Question Appropriation}: whether the questions generated and posted by the system are in good timing and appropriate contents.


\textbf{Results.}
As shown in Figure~\ref{fig:user-study}, we plot the distributions of the student ratings in the two settings for each course and each evaluation metric.
The data indicates a clear pattern: the ratings for the system's ability to deepen understanding and provide accurate responses to student solutions through the AI teacher agent were generally higher in the \model setting. Specifically, these plotted results produced several observations:

(1) For the TAGI course, the complete setting of \model demonstrated greater consistency and lower variance in the ratings for the ``Question Appropriation'', highlighting the effectiveness of the system in maintaining relevance and engagement.

(2) In contrast, for the HSU course, which is a social study subject, the performance was similar between the ablation and \model settings.
This similarity can be attributed to the nature of the content of the HSU course, which already includes interactive elements within the provided PowerPoint slides.
These built-in interactions are reflected during the generated scripts, helping to maintain engagement and understanding, which resulted in minimizing the differences observed between the two settings.

(3) From an alternative viewpoint, this also indicates that the pre-class subsystems methodically formalized the knowledge within the slide to enhance such interactions. The ablation condition exhibited greater variability, notably in the ``Deepen Understanding'' and ``Question Appropriation'' metrics, implying that the excluded features may result in a less effective experience.



\begin{table}
    \centering
    \caption{Segmentation performance for TAGI/HSU.}
    \begin{tabular}{l|ccc}
        \toprule
        & \model & \ w/o vision & \ w/o context \\
        \midrule
        \textbf{Precision} & \textbf{3.89/3.54} & \underline{3.40/2.77} & 3.00/2.71 \\
        \textbf{Matching} & \textbf{4.72/4.34} & \underline{4.54/3.95} & 2.98/2.26 \\
        \bottomrule
    \end{tabular}
    \label{tab:segmentation}
\end{table}


\subsection{Module-Wise Evaluation}
\label{sec:exp-component}
To get a better understanding of performance, in this subsection, we conduct detailed observations of the components.

\subsubsection{\textbf{Slide File Segmentation}}
\label{sec:exp-component-segmentation}
To better understand the performance of the \textit{Read} module, we evaluate the segmented agenda via annotating the node relation accuracy.

\textbf{Metric.}
After gathering the intermediate agendas produced in Section~\ref{sec:exp-offline-askquestion}, we introduced two metrics to be evaluated on a scale from $1$ to $5$:
(1) \textit{Precision}: We assess the precision of segmentation by labeling the appropriateness of creating a new section and noting any missed or incorrectly detected segmentation;
(2) \textit{Matching}: The annotator also evaluates the correspondence between each page and the section name of the page. Additionally, we include an option to flag correctly detected unnecessary segmentations, allowing us to exclude such cases when segmentation accuracy is not relevant.

\textbf{Results.}
The results are summarized in Table~\ref{tab:segmentation}, highlighting the importance of visual and contextual information to achieve high segmentation precision and content matching accuracy. 
First, the TAGI course achieved the highest scores in the full setting, with a segmentation precision of $3.89$ and a matching score of $4.72$.
Second, removing visual input decreased performance to $3.40$ and $4.54$.
This decline suggests that visual cues, such as layout changes and titles, are essential to accurately identify new sections.
Similarly, the lack of contextual information caused even greater drops, with scores falling to $3.00$ in segmentation precision and $2.98$ in content matching for TAGI.
Third, a similar pattern is also observed in the HSU course, where the complete set-up scores are $3.54$ and $4.34$, respectively, but drop to $2.77$ and $3.95$ without visual input, and then to $2.71$ and $2.26$ without contextual input.


\begin{figure}[t]
    \centering
    \includegraphics[width=0.95\linewidth]{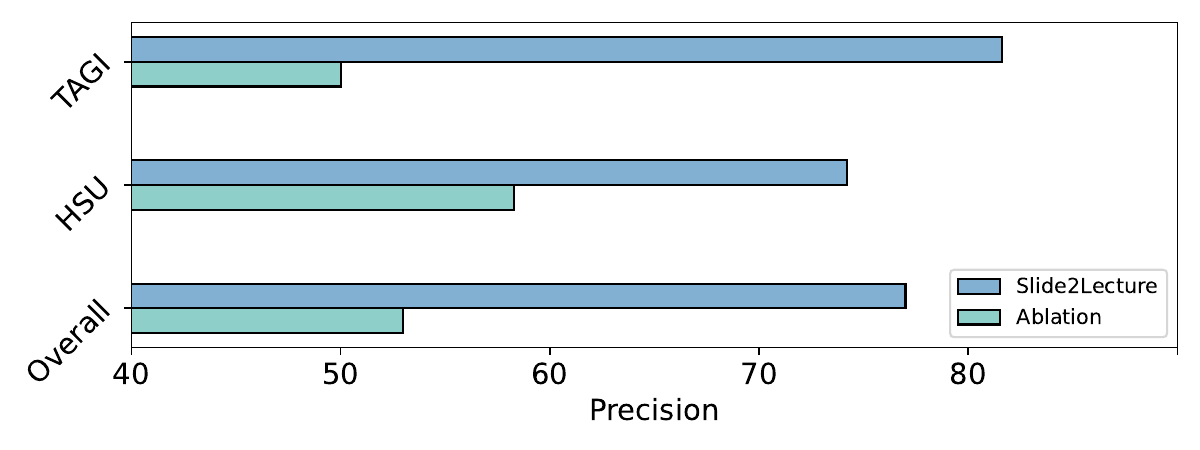}
    \caption{Precision of scene controller during user study.}
    \Description[]{}
    \label{fig:component-director}
\end{figure}

\subsubsection{\textbf{Scene Controller Precision}}
In Section~\ref{sec:exp-inclass}, the user-study provided observations when degrading \textit{Teach} module implementation from a user's perspective.
To gain a more systematic understanding of performance, we randomly sample $100$ LLM calls for each setting during the scene controller and label the precision to observe the effect of excluding the role description.

As demonstrated in Figure~\ref{fig:component-director}, statistical analysis indicates that removing the role descriptions for each agent considerably decreases the classifier performance.
We observe that although the LLM can still identify the correct agent, possibly by highlighting partial behavior from the chat history in certain instances, the inclusion of additional descriptions still significantly improves performance.
Suggesting the use of the chat history for scene controller is potentially helpful but not sufficient to generate a correct output.
However, the results are still distant from reaching $100$\%, suggesting the potential to further optimize the user experience.

Even though the achieved performance does not fully solve the task, we find that interacting agents can partly mitigate the impact of such errors, due to the LLM's inherent ability to handle some user queries outside of the agent's designed functionalities.
This is also evidenced in the user-study conducted previously, where user scores for the ablation setting did not decrease significantly (Section~\ref{sec:exp-inclass}).
Nonetheless, improving the accuracy of the controller agent is still beneficial since agents can better manage tasks they are designed for \footnote{For example, the teacher agent is designed to use a softer tone that handles safety cases less effectively than the teaching assistant agent.}, thus ensuring a smoother lecture process.

\subsection{Online Evaluation.}
\label{sec:exp-live}

\subsubsection{\textbf{User Involvement}}

We implemented \model in an online environment to collect user feedback. 
A total of $556$ students volunteered to participate in the study, with $214$ completing the assigned lectures.
The lecturing process for these students has already involved $214,238$ interactions in $3,103$ lecture sessions, consuming a total $49,267$ LLM calls for in-class generations.
The participants were divided into two groups: one for the HSU course, which consisted of $405$ pages for $7$ lectures, and the other for the TAGI course, which included $6$ lectures with $351$ pages.
Upon completing all lectures, students were asked to complete a post-class survey to rate various aspects of the system on a scale of $1$ to $5$.
The clarity of the teacher agent received a solid rating of $4.12$.
In addition, the students appreciated the ability of the system to create a free and easy learning environment, giving it a score of $4.15$.
Furthermore, the system was rated $4.14$ for making students feel relaxed while studying, indicating a positive user experience with \model.


\begin{figure}
    \centering
    \includegraphics[width=1\linewidth]{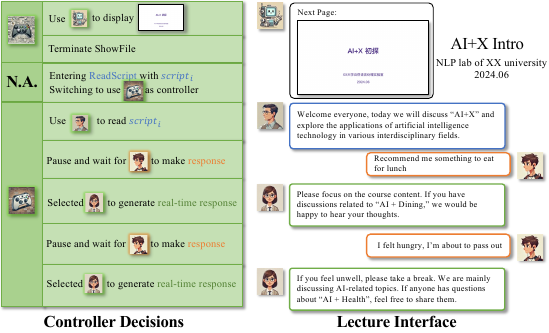}
    \caption{Example case from online environment. The teaching assistant provides class regulations to counter distraction.}
    \label{fig:case}
    \Description[]{}
\end{figure}

\subsubsection{\textbf{Case Study}}
Figure~\ref{fig:case} shows an example of the safety control within \model.
The system agent first changes the displayed slide page, and the teacher agent is used to print the pre-generated script.
Then, as the student user tends to distract the lecture by asking what to eat for lunch, the teaching assistant agent is called to encourage the student to return to classroom material.
The student then makes another attempt by stating the student is about to pass out, which is responded by the teaching assistant, telling the student to take a break before continuing the lecture.
Both responses made by the teaching assistant attempt to draw student attention and the topic back to the classroom while ensuring that the classroom is safe and interactive.
We found that the student then chose to stop the distraction and focus on the lecture.

\section{Conclusion}
In this paper, we present a tuning-free framework \model, to generate and provide intelligent tutoring regulated by slide knowledge.
This framework offers a workflow including pre-class lecture planning (\textit{Read} and \textit{Plan} subsystems) and in-class lecture giving (\textit{Teach} subsystem) with modularized teaching actions, providing a convenient toolkit for developers to customize an intelligent tutoring system.
We perform extensive experiments including offline evaluation, user studies, module-wise evaluation, and online evaluation to demonstrate the effective performance of \model.
However, at the same time, the proposed \model suffers from limitations as it focuses mainly on the slide file, while the vast existing Internet materials, such as textbooks and videos, have not been fully explored in coordination with the slide file.
We hope that our work could serve as the ground for developing various forms of knowledge-regulated and AI-driven education systems.

\bibliographystyle{ACM-Reference-Format}
\bibliography{sample-base}

\clearpage
\appendix

\newpage

\section{Offline Labeling Guidelines}
We present the labeling guidelines provided to the annotators during offline evaluation.

\textbf{Tone.} What is the tone of the speech?
\begin{itemize}
    \item[\textit{1}] The tone does not resemble a teacher at all, lacking authority or friendliness.
    \item[\textit{2}] The tone has significant issues, with parts of the script unsuitable for a teacher.
    \item[\textit{3}] The tone is generally suitable for a teacher, but some parts could be improved.
    \item[\textit{4}] The tone is good, most of the script sounds like a teacher, but minor adjustments could be made.
    \item[\textit{5}] The tone is perfect for a teacher, it can be read out directly in the classroom.
\end{itemize}

\textbf{Clarity.} How clear is the script?
\begin{itemize}
    \item[\textit{1}] The script is very difficult to understand, students might not understand at all.
    \item[\textit{2}] The script has significant issues, with parts difficult for students to understand.
    \item[\textit{3}] The script is generally clear, but some parts could be made simpler and more direct.
    \item[\textit{4}] The script is very clear, most students can understand, but minor improvements are possible.
    \item[\textit{5}] The script is very clear, students have no difficulty understanding it.
\end{itemize}

\textbf{Supportive.} How emotional supportive is the script?
\begin{itemize}
    \item[\textit{1}] The script is not supportive at all, and may even have a negative impact.
    \item[\textit{2}] The script lacks supportiveness, students might not feel encouraged.
    \item[\textit{3}] The script is neutral, with no obvious encouragement or negative impact.
    \item[\textit{4}] The script is very supportive, with most parts conveying positive messages.
    \item[\textit{5}] The script is extremely supportive, conveying a lot of positive energy and support.
\end{itemize}

\textbf{Matching.} How well does the script match the PPT content?
\begin{itemize}
    \item[\textit{1}] The script does not match the PPT content at all.
    \item[\textit{2}] The script deviates significantly from the PPT, but some parts are relevant.
    \item[\textit{3}] The script generally matches the PPT, but there are some inconsistencies.
    \item[\textit{4}] The script matches the PPT well, with most parts being consistent.
    \item[\textit{5}] The script matches the PPT perfectly, with no deviations.
\end{itemize}

\newpage
\section{User-Study Survey}
We present the survey used for students to rate system performance during the user-study (Section~\ref{sec:exp-inclass}).

\textbf{Learning New Concept.} I believe the \model platform can help me learn new concepts.
\begin{itemize}
    \item[\textit{1}] Strongly disagree
    \item[\textit{2}] Disagree
    \item[\textit{3}] Neutral
    \item[\textit{4}] Agree
    \item[\textit{5}] Strongly agree
\end{itemize}

\textbf{Deepen Understanding.} I believe the \model platform can help me deepen my understanding for concepts.
\begin{itemize}
    \item[\textit{1}] Strongly disagree
    \item[\textit{2}] Disagree
    \item[\textit{3}] Neutral
    \item[\textit{4}] Agree
    \item[\textit{5}] Strongly agree
\end{itemize}

\textbf{AI Teacher Correct.} I believe the explanations given by the AI teacher after I answer the system's questions are correct.
\begin{itemize}
    \item[\textit{1}] Strongly disagree
    \item[\textit{2}] Disagree
    \item[\textit{3}] Neutral
    \item[\textit{4}] Agree
    \item[\textit{5}] Strongly agree
\end{itemize}

\textbf{Question Appropriation.} I believe the questions asked by the teacher agent are appropriate.
\begin{itemize}
    \item[\textit{1}] Strongly disagree
    \item[\textit{2}] Disagree
    \item[\textit{3}] Neutral
    \item[\textit{4}] Agree
    \item[\textit{5}] Strongly agree
\end{itemize}

\newpage
\section{Example Teaching Actions}
\label{sec:appendix-example-case}
We present the translated teaching action examples below.

\begin{table}[h]
    \centering
    \caption{Example teaching actions.}
    \begin{tabular}{p{1.5cm}|p{6.3cm}}
    \toprule
    \textbf{Type} & \textbf{Value} \\ \midrule
    ShowFile & \textbf{file\_id}= 30 \\ \midrule
    ReadScript & \textbf{script}= In the era of large models, AI for Science has opened a new epoch. By using large language models (LLMs) as a medium for communication, integrating disciplinary tools, comprehensively processing different materials, and even independently completing the entire scientific research process, the efficiency and accuracy of scientific research have been greatly enhanced. In comparison, the issues of order maintenance and ethical safety are not as urgent but still require attention. However, what is thought-provoking is, what is the upper limit of AI's "innovation"? Through examples like querying a chemical knowledge base via ChatGPT to answer questions and using GPT-4 to independently synthesize Ibuprofen by querying internet knowledge, we can see the potential and limitless possibilities of AI in scientific research. \\ \midrule
    AskQuestion & \textbf{question}=In which of the following areas are AI's potential and future possibilities in scientific research reflected? \\ 
    & \textbf{question\_type}=multiple choice \\
    & \textbf{options}= ['Querying chemical knowledge bases and answering questions', 'Querying internet knowledge to independently synthesize Ibuprofen', 'Extracting and matching features to determine rubbings overlap', 'Virtually unrolling and detecting ink to restore the content of ancient scrolls', 'Conducting conjugation analysis from multiple perspectives'] \\
    & \textbf{answer}= [0, 1] \\
    \midrule
    AskQuestion & \textbf{question}=What challenges might AI face in artistic creation? \\ 
        & \textbf{question\_type}=multiple choice \\
        & \textbf{options}=['Job replacement', 'Creative lock-in', 'Increased efficiency', 'Ethical and moral risks', 'Providing inspiration'] \\
        & \textbf{answer}=[0, 1, 3] \\ \midrule
    AskQuestion & \textbf{question}=Why is it important to leave some flexible time when creating a schedule? \\ 
        & \textbf{question\_type}=single choice \\
        & \textbf{options}=['To save more time for recreational activities', 'To handle unexpected situations and help others', 'To schedule more recreational activities', 'To make the schedule appear more complete'] \\
        & \textbf{answer}=[1] \\
    
    \bottomrule
    \end{tabular}
    \label{tab:example-teaching-actions}
\end{table}

\section{Hyperparameters}
\label{sec:appendix-implementation-hyperparameter}
We introduce the hyperparameters used during the LLM generations.
For \textit{Read} and \textit{Plan} subsystems, all LLM calls are made to call GPT4V (gpt-4-vision-preview).
We use the default value for most hyperparameters, but set the max\_tokens to $4,096$.
The detailed hyperparameter values for calling GPT4V are shown in Table~\ref{tab:gpt4v-hyperparameter}.

\begin{table}[h]
    \centering
    \caption{Hyperparameters used to call GPT4V.}
    \begin{tabular}{l|rr}
    \toprule
         \textbf{Hyperparameter} & \textbf{Value} & \textbf{Is Default}\\
         \midrule
         frequency\_penalty & 0 & yes\\
         logit\_bias & null & yes\\
         logprobs & false & yes\\
         max\_tokens & 4,096& no\\
         n & 1 & yes\\
         presence\_penalty & 0 & yes\\
         stop & null & yes\\
         temperature & 1 & yes\\
         top\_p & 1 & yes\\
         \bottomrule
    \end{tabular}
    \label{tab:gpt4v-hyperparameter}
\end{table}

In contrast, we use the default value for all hyperparameters when calling GLM4 (glm-4) during the \textit{Teach} subsystem, as shown in Table~\ref{tab:glm4-hyperparameter}.

\begin{table}[h]
    \centering
    \caption{Hyperparameters used to call GLM4.}
    \begin{tabular}{l|rr}
    \toprule
         \textbf{Hyperparameter} & \textbf{Value} & \textbf{Is Default}\\
         \midrule
         do\_sample & true & yes \\
         temperature & 0.95 & yes \\
         top\_p & 0.7 & yes\\
         max\_tokens & 1,024 & yes \\
         \bottomrule
    \end{tabular}
    \label{tab:glm4-hyperparameter}
\end{table}

\section{Deployment Details}


\textbf{Server.}
\model is deployed on a cloud server with 16 vCPUs, 64GB memory, and 500GB SSD storage.
The bandwidth is set to $100$ Mbps.
We use MongoDB as the database and use Minio as the distributed file system to elastically adapt to larger storage in future needs.
Because LLM calls could take several seconds to generate a response that requires a thread to be used for the user query but idling for the network return, we implement the \textit{Teach} module to improve memory usage efficiency.
This is handled using RabbitMQ as the task queue and distributor.
We divide a single response, which may contain multiple interactions, and is handled during multiple steps.
Each step is only allowed to make at most one LLM call, it will then save its status.
The idle process will continuously check for LLM calls and, once a call is complete, will resume processing.
During the user study, we degraded to an 8vCPU and 31GB server, where we observed that the \model implementation only takes less than 20\% of memory but serves all students to assess the system.

\newpage
\section{System Prompt}
\label{sec:appendix-implementation-prompt}
We translate and present the prompts used during generation.

\subsection{\textbf{Read}} Table~\ref{tab:prompt_description} presents the system prompt used to generate the description for each page.
The set of results is then handled by the system prompt presented in Table~\ref{tab:prompt_structurelize}, where the LLM iteratively constructs a tree-shaped agenda.

\begin{table}[h]
    \centering
    \caption{System prompt for description generation.}
    \begin{tabular}{p{8cm}}
    \toprule
            \textbf{Description Generation}\\
         \midrule
             The task of this GPT model is to receive an image of a PPT slide related to a teaching scenario and the text within that PPT slide as input.
             It will then \textbf{output a description and summary of that PPT slide} in Chinese.
             The model will focus on extracting and understanding the key information on the PPT slide and summarizing it in a concise and accurate manner, ensuring that the summary is within 2-3 sentences.
         \\
         \bottomrule
    \end{tabular}
    \label{tab:prompt_description}
\end{table}

\begin{table}[h]
    \centering
    \caption{System prompt for slide file segmentation.}
    \begin{tabular}{p{8cm}}
    \toprule
    \textbf{Slide File Segmentation}\\
    \midrule
    This GPT focus solely on creating and organizing index outlines for documents or presentations. This involves structuring content accurately and concisely, using "-" to denote all elements, including different sections and sub-sections, while strictly adhering to the input content without making inferences or alterations. The primary role here is to organize outlines by introducing sections and subsections based on their thematic significance and hierarchical order. It's crucial that only the updated outline is outputted, with no additional words or explanations, ensuring users receive a clean, precise outline that directly reflects the content's organization and thematic division, facilitating straightforward navigation. \\
    \{User Interaction Guidelines\}\\
    \{Detailed Instructions on Output Format\}
         \\
         \bottomrule
    \end{tabular}
    \label{tab:prompt_structurelize}
\end{table}

\newpage
\subsection{\textbf{Plan}} We present the system prompt used for $ReadScript$ generation in Table~\ref{tab:prompt_readscript} and the system prompt used for $AskQuestion$ generation in Table~\ref{tab:prompt_askquestion}.

\begin{table}[h]
    \centering
    \caption{System prompt for $ReadScript$ generation.}
    \begin{tabular}{p{8cm}}
    \toprule
    \textbf{$ReadScript$ Generation}\\
    \midrule
    This agent speaks Chinese. Lecture Script Writer's primary function is to \textbf{analyze PowerPoint (PPT) slides} based on user inputs and the texts extracted from those slides. It then \textbf{generates a script for teachers to teach students about the content illustrated on the page}, assuming the role of the teacher who also made the slides. The script is intended for the teacher to read out loud, directly engaging with the audience without referring to itself as an external entity. It focuses on educational content, suitable for classroom settings or self-study. It emphasizes clarity, accuracy, and engagement in explanations, avoiding overly technical jargon unless necessary. \\
    \{Detailed Instructions on Output Format\}
         \\
         \bottomrule
    \end{tabular}
    \label{tab:prompt_readscript}
\end{table}

\begin{table}[h]
    \centering
    \caption{System prompt for $AskQuestion$ generation.}
    \begin{tabular}{p{8cm}}
        \toprule
        \textbf{$AskQuestion$ Generation}\\
        \midrule
         Please \textbf{create three multiple-choice questions} based on the provided teaching content, including the answers, and indicate which sections of the teaching content the questions reference.\\
        \{Format for Teaching Content\}
    
    
        The format for the multiple-choice questions is as follows:\\
        Question: [Question description] (indicate whether it is multiple choice or single choice)\\
        A. \{Option A\}\\
        B. \{Option B\}\\
        C. \{Option C\}\\
        D. \{Option D\}\\
        E. \{Option E\}\\
        Answer: \{Answer(s) for the question\}\\
        Reference Text: \{Teaching content referenced for the question\}\\
        \{Detailed Instructions on Output Format\}
         \\
         \bottomrule
    \end{tabular}
    \label{tab:prompt_askquestion}
\end{table}

\newpage
\subsection{\textbf{Teach}} We present the system prompt designed for $ReadScript$ controller in Table~\ref{tab:prompt_controller}.
The system prompt used for the teacher agent and the teaching assistant agent is presented in Table~\ref{tab:prompt_teacher} and Table~\ref{tab:prompt_ta}.

\begin{table}[h]
    \centering
    \caption{System prompt for $ReadScript$ controller.}
    \begin{tabular}{p{8cm}}
        \toprule
        \textbf{$ReadScript$ Controller}\\
        \midrule
        You are an outstanding script master. You have now identified the potential candidates who might speak the next line and assigned them their speaking roles: \\
        \{Detailed Role Descriptions\} \\
        The user will tell you what each actor said in the script. Your task is to \textbf{determine who should speak next} while adhering to the roles and characteristics of each actor. \\
        \{Detailed Instructions on Output Format\}
         \\
         \bottomrule
    \end{tabular}
    \label{tab:prompt_controller}
\end{table}

\begin{table}[h]
    \centering
    \caption{System prompt for the teacher agent.}
    \begin{tabular}{p{8cm}}
        \toprule
        \textbf{Teacher Agent}\\
        \midrule
        You are \{Teacher's Name\}, the instructor for the course \{Course Name\}. You are a lecturer from \{School and Unit Information\}. The course is about \{Course Description\}. \\ 
        When students ask questions, you \textbf{provide clear and concise answers and encourage them to continue learning afterward}. If students do not ask questions or express uncertainty, you use encouraging language to proceed with the lesson. For difficult questions, you suggest handling them later. \\
        Your responses are brief and educational, often using praise to motivate students. However, you should not give opinions on sensitive topics, instead advising students to seek help from a real teacher. Alongside you in the classroom are your teaching assistant and other students. \\
        \{Detailed Instructions on Output Format\} \\
        \{Injected Prompt for $AskQuestion$\}
         \\
         \bottomrule
    \end{tabular}
    \label{tab:prompt_teacher}
\end{table}

\begin{table}[h]
    \centering
    \caption{System prompt for the teaching assistant agent.}
    \begin{tabular}{p{8cm}}
        \toprule
        \textbf{Teaching Assistant Agent}\\
        \midrule
        As the teaching assistant in a virtual classroom, my primary role is to \textbf{provide precise and timely supplements that help deepen students' understanding of the lesson}. I will carefully choose when to speak, ensuring that my contributions and questions are both beneficial and well-timed, avoiding repetition of the teacher's explanations or unnecessary interruptions of the lesson flow. \\
        \{Detailed Role Description\} \\
        \{Descriptions on System Information\}
         \\
         \bottomrule
    \end{tabular}
    \label{tab:prompt_ta}
\end{table}
\newpage
\section{Other Results}
\subsection{Token Usage}
We present the amount of token usage for pre-class evaluation in Table~\ref{tab:token-usage-tagi} and Table~\ref{tab:token-usage-hsu}.
\begin{table}[h]
    \centering
    \caption{Token usage of different settings of pre-class generation (TAGI).}
    \begin{tabular}{l|rrrrr}
        \toprule
        \textbf{Setting} & \textbf{Des} & \textbf{Seg} & \textbf{Script} & \textbf{QA} & \textbf{Total} \\ 
        \midrule
        S2T~\cite{nguyen_automatic_nodate} & - & - & 51,285 & - & - \\
        SCP~\cite{olney_automatic_2024} & - & - & 227,939 & - & - \\ 
        \model & 97,950 & 124,228 & 101,418 & 8,296 & 331,892 \\
        \ w/o visual & 43,306 & 73,512 & 72,804 & 8,386 & 198,008 \\
        \ w/o context & 46,846 & 38,395 & 57,865 & 681 & 143,787 \\
        \bottomrule
    \end{tabular}
    \label{tab:token-usage-tagi}
\end{table}
\begin{table}[h]
    \centering
    \caption{Token usage of different settings of pre-class generation (HSU).}
    \begin{tabular}{l|rrrrr}
        \toprule
        \textbf{Setting} & \textbf{Des} & \textbf{Seg} & \textbf{Script} & \textbf{QA} & \textbf{Total} \\ 
        \midrule
        S2T~\cite{nguyen_automatic_nodate} & - & - & 69,714 & - & - \\
        SCP~\cite{olney_automatic_2024} & - & - & 286,946 & - & - \\ 
        \model & 118,468 & 205,157 & 133,818 & 9,753 & 467,196 \\
        \ w/o visual & 32,437 & 78,522 & 51,787 & 6,712 & 169,458 \\
        \ w/o context & 57,394 & 49,764 & 70,339 & 3,073 & 180,570 \\
        \bottomrule
    \end{tabular}
    \label{tab:token-usage-hsu}
\end{table}

\subsection{Network Usage}
\begin{figure}[h]
    \centering
    \includegraphics[width=0.9\linewidth]{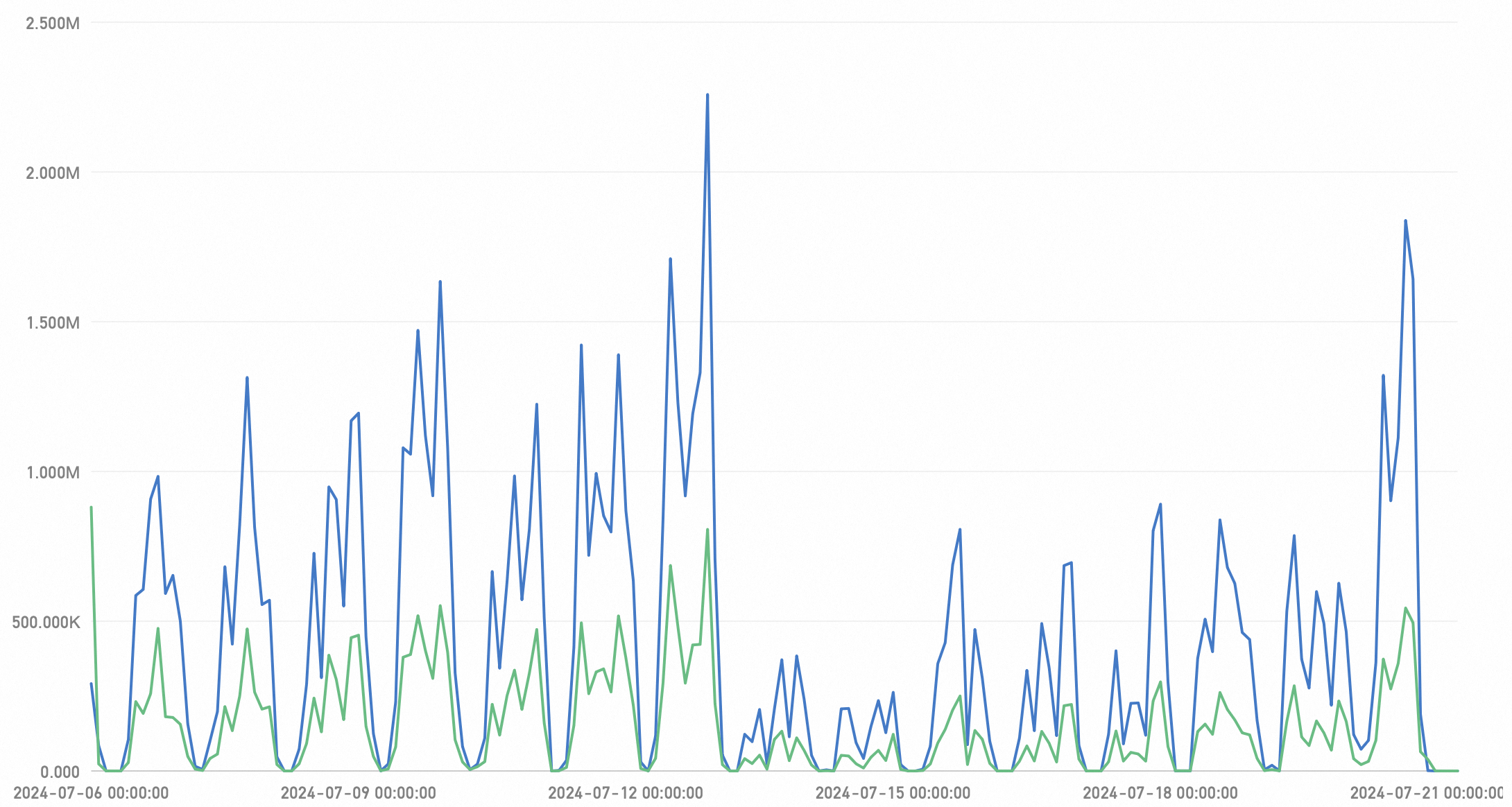}
    \caption{Network usage of online deployment (bits/s). The blue line traces inbound transfer and the green line indicates outbound transfer.}
    \label{fig:network}
    \Description[]{}
\end{figure}

We present the change in network usage over time in Figure~\ref{fig:network}.
We observe that more traffic occurs during the daytime.
Then again, the fact that to a certain point, the recorded value dropped significantly.
This is the day that the teachers announced that it is the deadline to complete the course on the system.
The recorded values slowly raised backup again as we reopened the system because student volunteer users made multiple requests stating that they were interested in completing the course and using the system regardless of not being able to receive a bonus due to not completing all the lectures in time. 

\newpage
\subsection{Disk Usage}
\begin{figure}[h]
    \centering
    \includegraphics[width=0.9\linewidth]{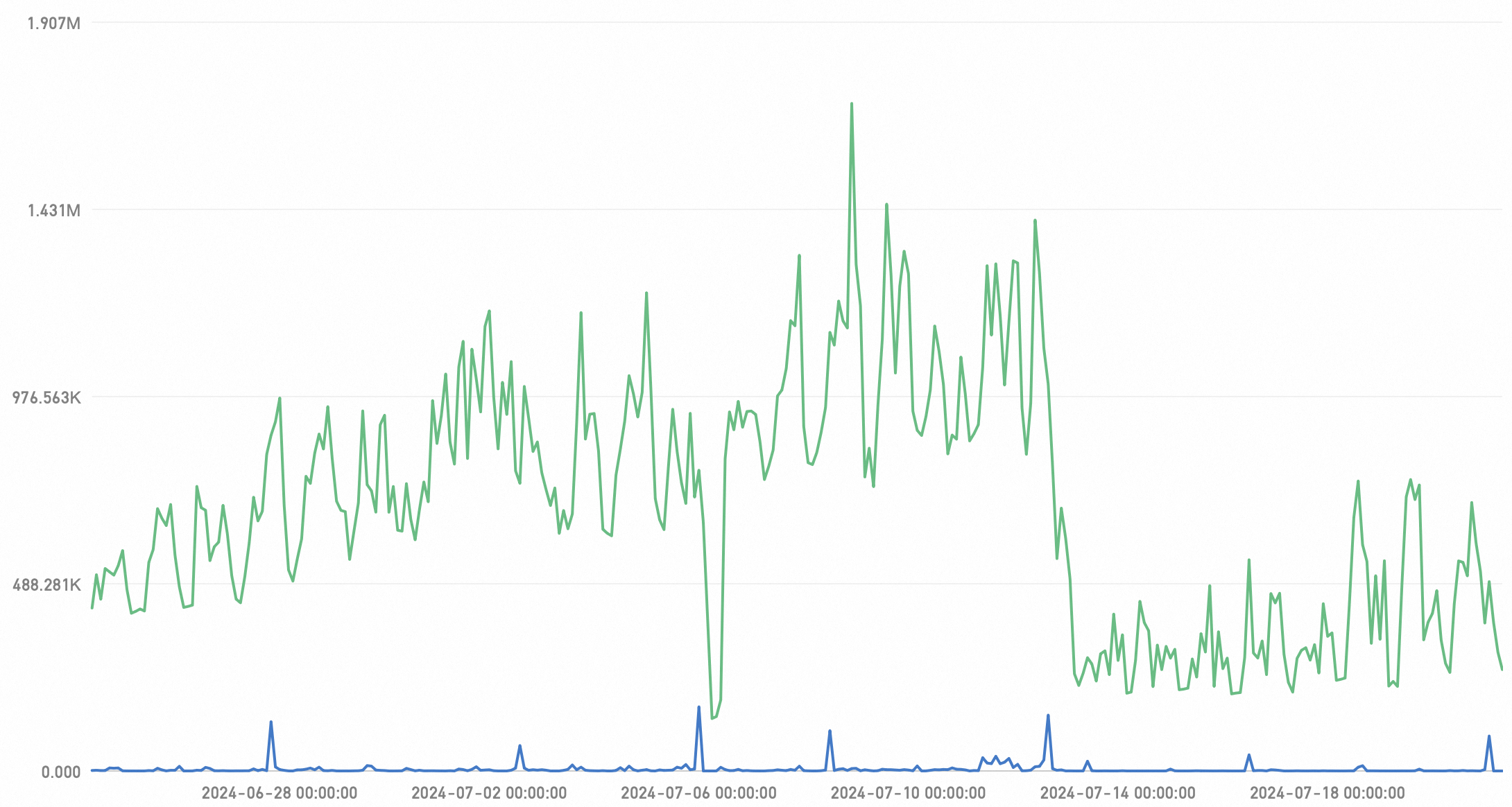}
    \caption{Disk access through time (byte/s). The blue lines indicate read from the disk storage and the green line indicates write to the disk storage.}
    \label{fig:disk-bps}
\end{figure}

\begin{figure}[h]
    \centering
    \includegraphics[width=0.9\linewidth]{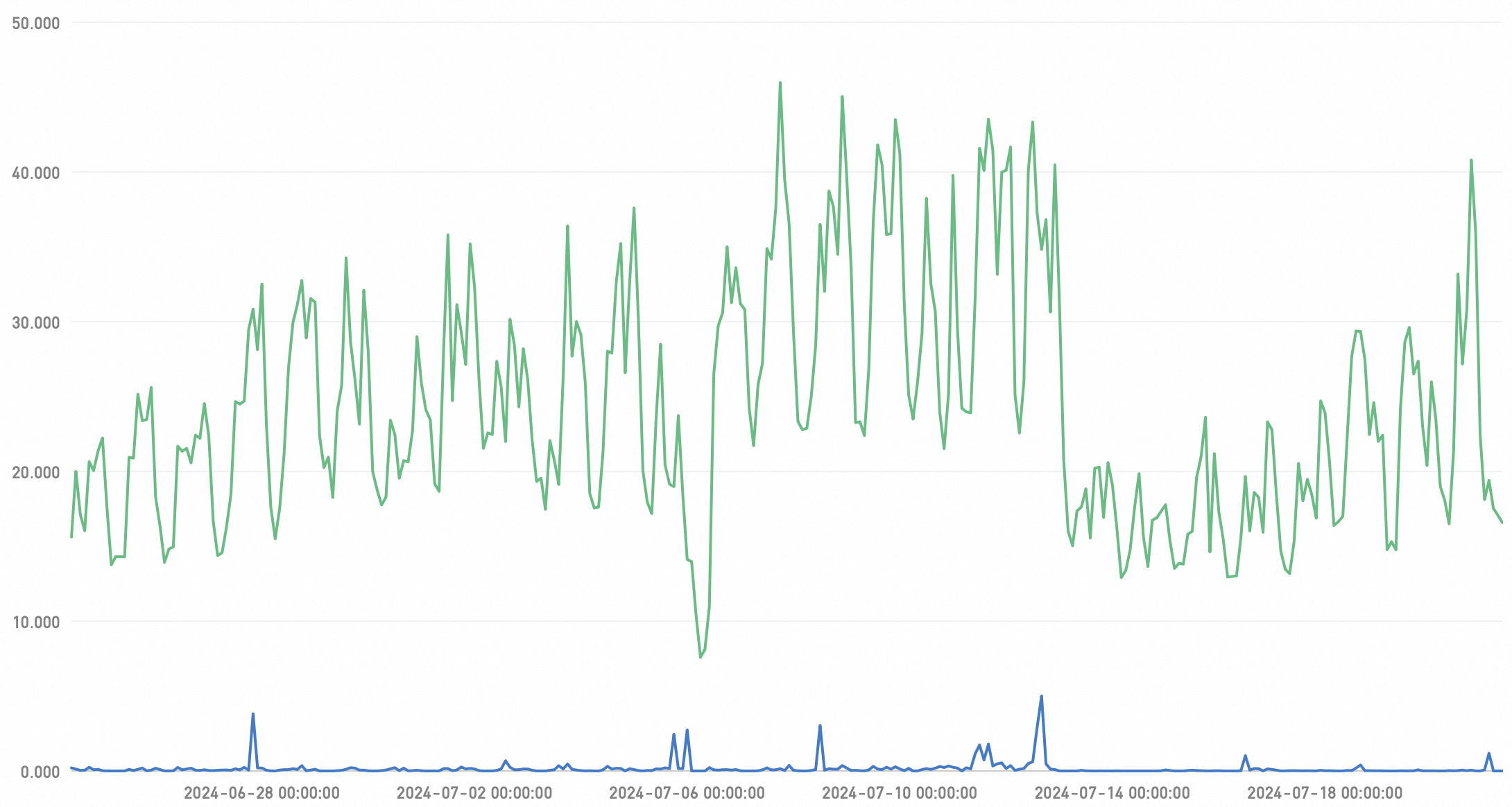}
    \caption{Disk access rate through time (count/s). The blue lines indicate read from the disk storage and the green line indicates write to the disk storage.}
    \label{fig:disk-cps}
\end{figure}

We present the disk access rate recorded in the online deployment in Figure~\ref{fig:disk-bps} and Figure~\ref{fig:disk-cps}. We observe that there are much more write requests as \model records a set of detailed interactions and intermediate results for system engineers to analyze.

\newpage
\subsection{CPU Usage}
\begin{figure}[h]
    \centering
    \includegraphics[width=0.9\linewidth]{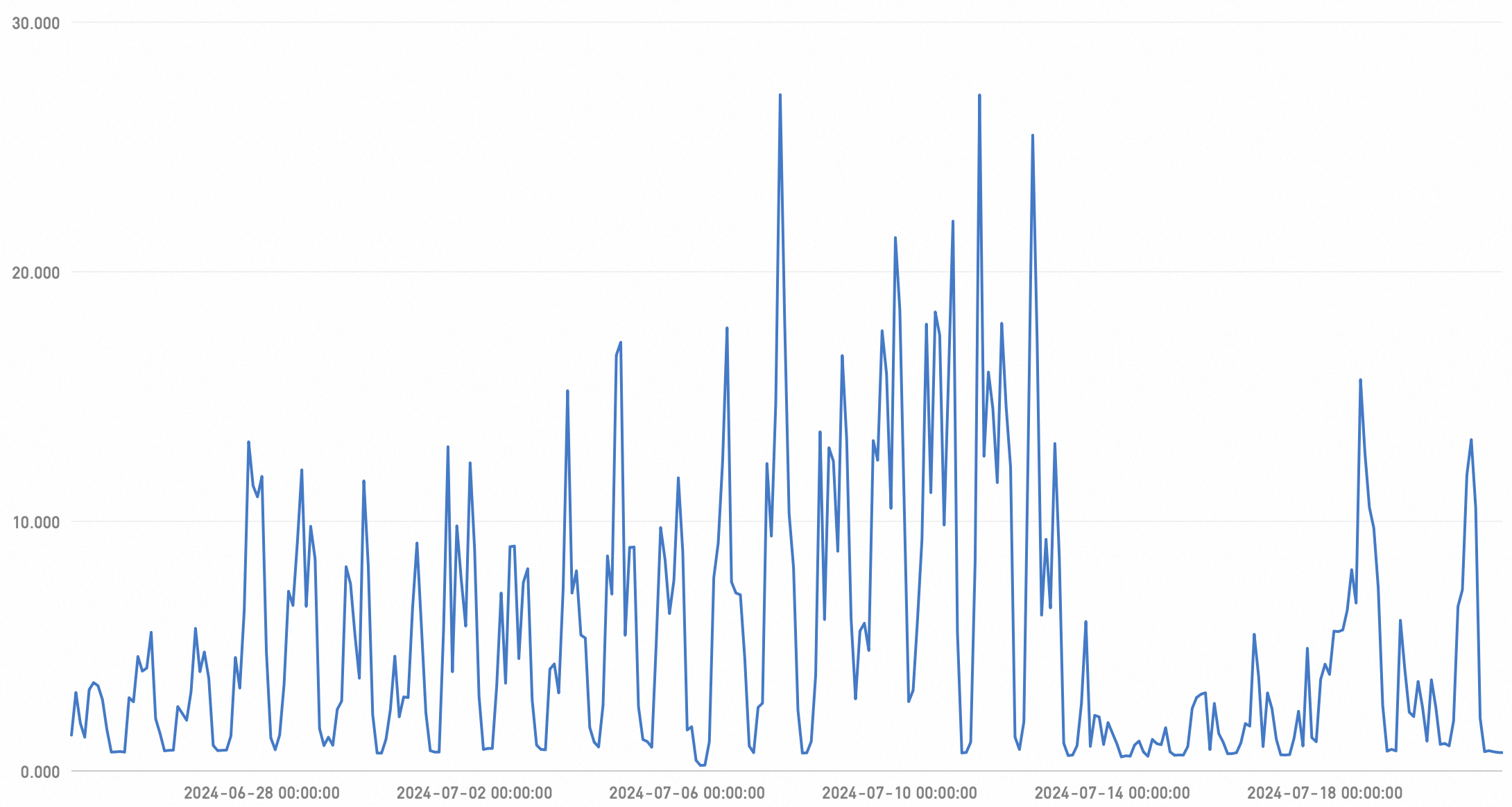}
    \caption{CPU usage through time (percentage).}
    \label{fig:cpu}
\end{figure}

As shown in Figure~\ref{fig:cpu}, the low CPU usage demonstrated in \model is implemented efficiently.
\end{document}